%% file: main.tex
\documentclass[runningheads]{llncs}
\usepackage{graphicx}
\usepackage{epsfig}
\usepackage{amsmath}
\usepackage{amssymb}
\usepackage{multirow}
\usepackage{floatrow}
\usepackage{adjustbox}
\usepackage{tabularx}
\usepackage{booktabs}
\usepackage{flushend}
\usepackage[table, dvipsnames]{xcolor}
\usepackage{caption}
\usepackage{comment}
\usepackage[colorlinks=true, linkcolor=blue, urlcolor=blue, citecolor = green]{hyperref}%
\usepackage{floatrow}
\usepackage{multirow}
\usepackage[misc,geometry]{ifsym}

\newcommand\sanjana[1]{\textcolor{red}{[S: #1]}}
\begin{document}
\title{Towards Boosting the Accuracy of Non-Latin Scene Text Recognition}
%
%
\author{Sanjana Gunna(\Letter)\orcidID{0000-0003-3332-8355}, Rohit Saluja \orcidID{0000-0002-0773-3480},  \and \\ C. V. Jawahar \orcidID{0000-0001-6767-7057}}

%
\authorrunning{Gunna et al.}
%
\institute{Centre for Vision Information Technology \\
International Institute of Information Technology, Hyderabad - 500032, INDIA \\
\url{https://github.com/firesans/NonLatinPhotoOCR} \\
\email{\{sanjana.gunna,rohit.saluja\}@research.iiit.ac.in}, \email{jawahar@iiit.ac.in}}
%
\maketitle              
\begin{abstract}
Scene-text recognition is remarkably better in Latin languages than the non-Latin languages due to several factors like multiple fonts, simplistic vocabulary statistics, updated data generation tools, and writing systems. This paper examines the possible reasons for low accuracy by comparing English datasets with non-Latin languages. We compare various features like the size (width and height) of the word images and word length statistics. Over the last decade, generating synthetic datasets with powerful deep learning techniques has tremendously improved scene-text recognition. Several controlled experiments are performed on English, by varying the number of (i) fonts to create the synthetic data and (ii) created word images. We discover that these factors are critical for the scene-text recognition systems. The English synthetic datasets utilize over 1400 fonts while Arabic and other non-Latin datasets utilize less than 100 fonts for data generation. Since some of these languages are a part of different regions, we garner additional fonts through a region-based search to improve the scene-text recognition models in Arabic and Devanagari. We improve the Word Recognition Rates (WRRs) on Arabic MLT-17 and MLT-19 datasets by $24.54\%$ and $2.32\%$ compared to previous works or baselines. We achieve WRR gains of $7.88\%$ and $3.72\%$ for IIIT-ILST and MLT-19 Devanagari datasets. 

\end{abstract}
\input{1Intro}
\input{2Dataset}
\input{3Models}
\input{5Results}

\input{6Conclusions}
\bibliographystyle{splncs04}
\bibliography{paper}
\end{document}

%% file: 1Intro.tex
\begin{figure}[ht]
    \centering
    \includegraphics[width=0.9\linewidth]{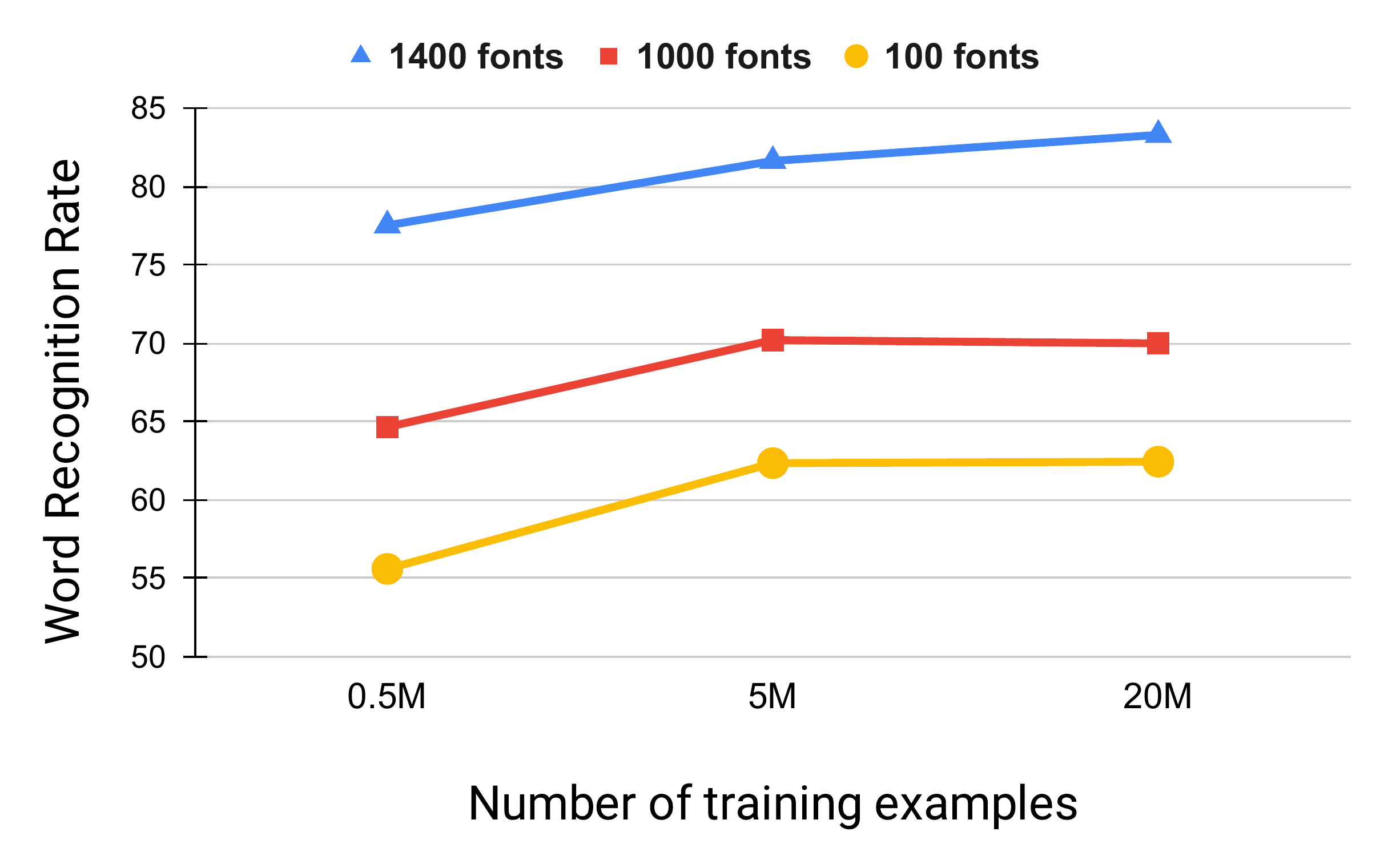}
    \caption{Comparing STAR-Net's performance on IIIT5K~\cite{MishraBMVC12} dataset when trained on synthetic data created using a varying number of fonts and training samples.}
    \label{fig:LATIN_WRR}
\end{figure}
\section{Introduction}\label{sec:Intro}
The task of scene-text recognition involves reading the text from natural images. It finds applications in aiding the visually impaired, extracting information for map services and geographical information systems by mining data from the street-view-like images~\cite{buvsta2018e2e}. The overall pipeline for scene-text recognition involves a text detection stage followed by a text recognition stage. Predicting the bounding boxes around word images is called text detection~\cite{huang2019mask}. The next step involves recognizing text from the cropped text images obtained from the labeled or the predicted bounding boxes~\cite{mathew2017benchmarking}. In this work, we focus on improving text recognition in non-Latin languages.
Multilingual text recognition has witnessed notable growth due to the impact of globalization leading to international and intercultural communication. Like English, the recognition algorithms proposed for Latin datasets have not successfully recorded similar accuracies on non-Latin datasets. Reading text from non-Latin images is challenging due to the distinct variation in the scripts used, writing systems,  scarcity of data, and fonts. In Fig.~\ref{fig:LATIN_WRR}, we illustrate the analysis of Word Recognition Rates (WRR) on the IIIT5K English dataset~\cite{MishraBMVC12} by varying the number of training samples and fonts used in the synthetic data. The training performed on STAR-Net~\cite{liu2016star} proves extending the number of fonts leads to better WRR gains than increasing training data.  We incorporate the new fonts found using region-based online search to generate synthetic data in Arabic and Devanagari. 
The motivation behind this work is described in Section~\ref{sec:datasets}.  The methodology to train the deep neural network  on the Arabic and Devanagari datasets is detailed in Section~\ref{sec:models}. The results and conclusions from this study are presented in Section~\ref{sec:results} and~\ref{sec:conclusions}, respectively. The contributions of this work are as follows:
\begin{enumerate}
    \item 
    We study the two parameters for synthetic datasets crucial to the performance of the reading models on the IIIT5K English dataset; i) the number of training examples and ii) the number of diverse fonts\footnote{We also investigated other reasons for low recognition rates in non-Latin languages, like comparing the size of word images of Latin and non-Latin real datasets but could not find any significant variations (or exciting differences). Moreover, we observe very high word recognition rates ($>90\%$) when we tested our non-Latin models on the held-out synthetic datasets, which shows that learning to read the non-Latin glyphs is trivial for the existing deep models. Refer~\url{https://github.com/firesans/STRforIndicLanguages} for more details.}.
    \item We share $55$ additional fonts in Arabic, and $97$ new fonts in Devanagari, which we found using a region-wise online search. These fonts were not used in the previous scene text recognition works.
    \item We apply our learnings to improve the state-of-the-art results of two non-Latin languages, Arabic, and Devanagari.
\end{enumerate}
\begin{table}[t]
\resizebox{\textwidth}{!}
{%
\centering
\begin{tabular}{lr}
\toprule
{\bf Language} & {\bf Datasets} \\
\midrule
Multilingual & IIIT-ILST-17 ($3K$ words, 3 languages), \\
& MLT-17 ($18 K$ scenes, $9$ languages), \\
&MLT-19 ($20K$ scenes, $10$ languages) \\
&OCR-on-the-go-19 (1000 scenes, 3 languages),\\ &CATALIST-21 (2322 scenes, 3 languages)\\
Arabic & ARASTEC-15 ($260$  signboards, hoardings, advertisements),  MLT-17,19\\
Chinese & RCTW-17 ($12 K$ scenes), ReCTS-25K-19 ($25 K$ signboards),\\
& CTW-19 ($32 K$ scenes), RRC-LSVT-19 ($450 K$ scenes), MLT-17,19\\
Korean & KAIST-11 ($2.4 K$ signboards, book covers, characters) , MLT-17,19\\
Japanese & DOST-16 ($32K$ images), MLT-17,19\\
English & SVT-10 ($350$ scenes), SVT-P-13 ($238$ scenes, $639$ words), \\
& IIIT5K-12 ($5K$ words),  IC11 ($485$ scenes, $1564$  words), \\
& IC13 (462 scenes), IC15 (1500 scenes), COCO-Text-16 ($63.7K$ scenes), \\ 
& CUTE80-14 (80 scenes), Total-Text-19 (2201 scenes), MLT-17,19\\ 

\bottomrule
\end{tabular}
}
\caption{Latin and non-Latin scene-text recognition datasets.} 
\label{tab:non_latin_data}
\end{table}
\section{Related Work}\label{sec:relatedwork}

Recently, there has been an increasing interest in scene-text recognition for a few widely spoken non-Latin languages around the globe, such as Arabic, Chinese, Devanagari, Japanese, Korean. Multi-lingual datasets have been introduced to tackle such languages due to their unique characteristics.  As shown in Table~\ref{tab:non_latin_data}, Mathew et al.~\cite{mathew2017benchmarking} release the IIIT-ILST Dataset containing around $1K$ images each three non-Latin languages. The MLT dataset from the ICDAR'17 RRC contains images from Arabic, Bangla, Chinese, English, French, German, Italian, Japanese, and Korean~\cite{nayef2017icdar2017}. The ICDAR'19 RRC builds MLT-19 on top of MLT-17 to containing text from Arabic, Bangla, Chinese, English, French, German, Italian, Japanese, Korean, and Devanagari~\cite{nayef2019icdar2019}. Recent OCR-on-the-go and CATALIST\footnote{\url{https://catalist-2021.github.io/}} datasets include around $1000$ and $2322$ annotated videos in Marathi, Hindi, and English~\cite{saluja2019ocrgo}. Arabic scene-text recognition datasets involve ARASTEC and MLT-17,19~\cite{tounsi2015arabic}. 
Chinese datasets cover RCTW, ReCTS-25k, CTW, and RRC-LSVT from ICDAR'19 Robust Reading Competition (RRC)~\cite{shi2017icdar2017,zhang2019icdar,yuan2019ctw,sun2019icdar}. Korean and Japanese scene-text recognition datasets involve KAIST and DOST~\cite{jung2011touch,iwamura2016downtown}. Different English datasets are listed in the last row of Table~\ref{tab:non_latin_data}~\cite{wang2010word,6126402,Shahab2011ICDAR2R,MishraBMVC12,Phan2013RecognizingTW,10.1109/ICDAR.2013.221,article1,veit2016coco,Chng2017TotalTextAC,nayef2017icdar2017,nayef2019icdar2019}.\\
Various models have been proposed for the task of scene-text recognition. 
Wang et al.~\cite{wang2011end} present an object recognition module that achieves competitive performance by training on ground truth lexicons without any explicit text detection stage. 
Shi et al.~\cite{shi2016end} propose a Convolutional  Recurrent  Neural  Network  (CRNN) architecture. It achieves remarkable performances in both lexicon-free and lexicon-based scene-text recognition tasks as is used by Mathew et al.~\cite{mathew2017benchmarking} for three non-Latin languages. 
Liu et al.~\cite{liu2016star} introduce Spatial Attention Residue Network (STAR-Net) with Spatial Transformer-based Attention Mechanism, which handles image distortions. 
 Shi et al.~\cite{shi2018aster} propose a segmentation-free Attention-based method for Text Recognition (ASTER). 
Mathew et al.~\cite{mathew2017benchmarking} achieves the Word Recognition Rates (WRRs) of $42.9\%$, $57.2\%$, and $73.4\%$ on $1K$ real images in Hindi, Telugu, and Malayalam, respectively. Bušta et al.~\cite{buvsta2018e2e} propose a CNN (and CTC) based method for text localization, script identification, and text recognition and is tested on $11$ languages (including Arabic) of MLT-17 dataset. The WRRs are above $65\%$ for Latin and Hangul and are below $47\%$ for the remaining languages ($46.2\%$ for Arabic). Therefore, we aim to improve non-Latin recognition models.


%% file: 2Dataset.tex
\section{Motivation and Datasets}~\label{sec:datasets}
This section explains the motivation behind our work. Here we also describe the datasets used for experiments on non-Latin scene text recognition.

{\bf Motivation:} 
To study the effect of fonts and training examples on scene-text recognition performance, we randomly sample $100$ and $1000$ fonts from the set of over $1400$ English fonts from previous works~\cite{Jaderberg14c,gupta2016synthetic}. For $1400$ fonts, we use the datasets available from earlier photo OCR works on synthetic dataset generation~\cite{Jaderberg14c,gupta2016synthetic}. For $100$ and $1000$ fonts, we generate synthetic images by following a simplified methodology proposed by Mathew et al.~\cite{mathew2017benchmarking}. Therefore, we create three different synthetic datasets. Moreover, we simultaneously experiment by varying the number of training samples from $0.5M$ to $5M$ to $20M$ samples. By changing the two parameters, we train our model (refer Section~\ref{sec:models}) on the above synthetic datasets and test them on the IIIT5K dataset. We observe that the Word Recognition Rate (WRR) of the dataset with around $20M$ samples and over $1400$ fonts achieves state-of-the-art accuracy on the IIIT5K dataset~\cite{liu2016star}. As shown in Fig. \ref{fig:LATIN_WRR}, the WRR of the model trained on $5M$ samples generated using over $1400$ fonts is very close to the recorded WRR ($20M$ samples). Moreover, models trained on $1400$ fonts outperform the models trained on $1000$ and $100$ fonts by a margin of $10\%$ because of improved (font) diversity and better but complex dataset generation methods. Also, in Fig.\ref{fig:LATIN_WRR}, as we increase the number of fonts from $1000$ to $1400$, the WRR gap between the models trained on $5M$ and $20M$ samples moderately improves (from $0\%$ to around $2\%$). Finally, this analysis highlights the importance of increasing the fonts in synthetic dataset generation and ultimately improving the scene-text recognition models.
\begin{table}[t]
\resizebox{0.65\textwidth}{!}
{%
\centering
\begin{tabular}{lcccrcr}
\toprule
{\bf Language} & {\bf \# Images} & ~ & {\bf $\mu$, $\sigma$ word length} & {\bf \# Fonts} \\
\midrule
English & 17.5M & ~ & 5.12, 2.99 & $>$1400\\
Arabic & 5M & ~ & 6.39, 2.26 & 140\\
Devanagari & 5M & ~ & 8.73, 3.10& 194\\
\bottomrule
\end{tabular}
}
\caption{Synthetic Data Statistics. $\mu$, $\sigma$ represent mean, standard deviation.}
\label{tab:synth_data}
\end{table}
\begin{figure}[t]
    \centering
    \includegraphics[trim=20 240 23 130, clip,width=\linewidth]{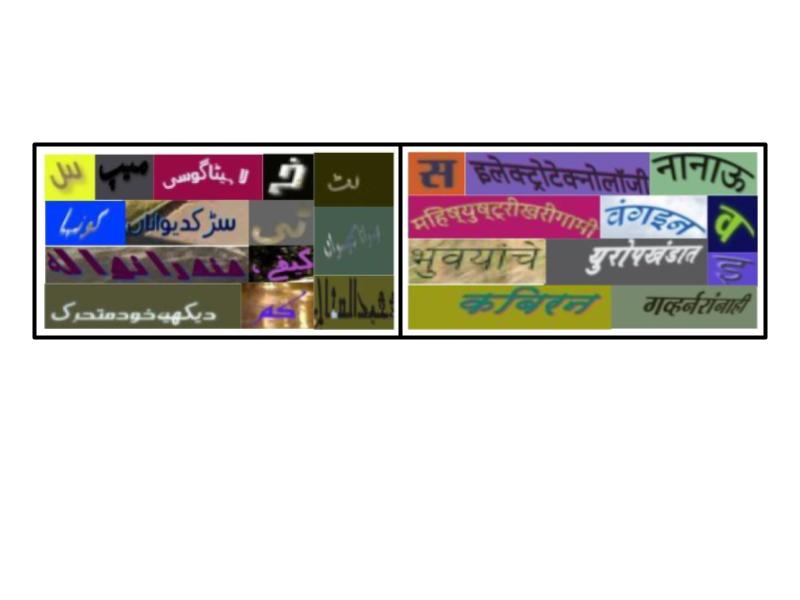}%
    \caption{Synthetic word images in Arabic and Devanagari.}
    \label{fig:sample_synth_images}
\end{figure}

{\bf Datasets:} 
As shown in Table~\ref{tab:synth_data}, we generate over $17M$ word images in English, $5M$ word images each in Arabic, and Devanagari, using the tools provided by Mathew et al.~\cite{mathew2017benchmarking}. We use $140$ and $194$ fonts for Arabic and Devanagari, respectively. Previous works use $97$ fonts and $85$ fonts for these languages~\cite{mathew2017benchmarking,buvsta2017deep}. Since the two languages are spoken in different regions, we found $55$ additional fonts in Arabic and $97$ new fonts in Devanagari using the region-wise online search.\footnote{Additional fonts we found using region-based online search are available at: ~\url{www.sanskritdocuments.org/}, ~\url{www.tinyurl.com/n84kspbx}, ~\url{www.tinyurl.com/7uz2fknu}, ~\url{www.ctan.org/tex-archive/fonts/shobhika?lang=en}, ~\url{www.hindi-fonts.com/}, ~\url{www.fontsc.com/font/tag/arabic}, more fonts are shared on~\url{https://github.com/firesans/NonLatinPhotoOCR}}. We use the additional fonts obtained by region-wise online search, which we will share with this work. As we will see in Section~\ref{sec:results}, we also perform some of our experiments with these fonts. 
Sample images of our synthetic data are shown in Fig.~\ref{fig:sample_synth_images}. As shown in Table~\ref{tab:synth_data}, English has the lowest average word length among the languages mentioned, while Arabic and Devanagari have comparable average word lengths. Please note that we use over $1400$ fonts for English, whereas the number of diverse fonts available for the non-Latin languages is relatively low. We run our models on Arabic and Devnagari test sets from MLT-17, IIIT-ILST, and MLT-19 datasets\footnote{We could not obtain the ARASTEC dataset we discussed in the previous section.}. The results are summarized in Section~\ref{sec:results}.

%% file: 3Models.tex
\begin{figure}[t]
    \centering
    \includegraphics[trim=0 320 0 10, clip,width=\linewidth]{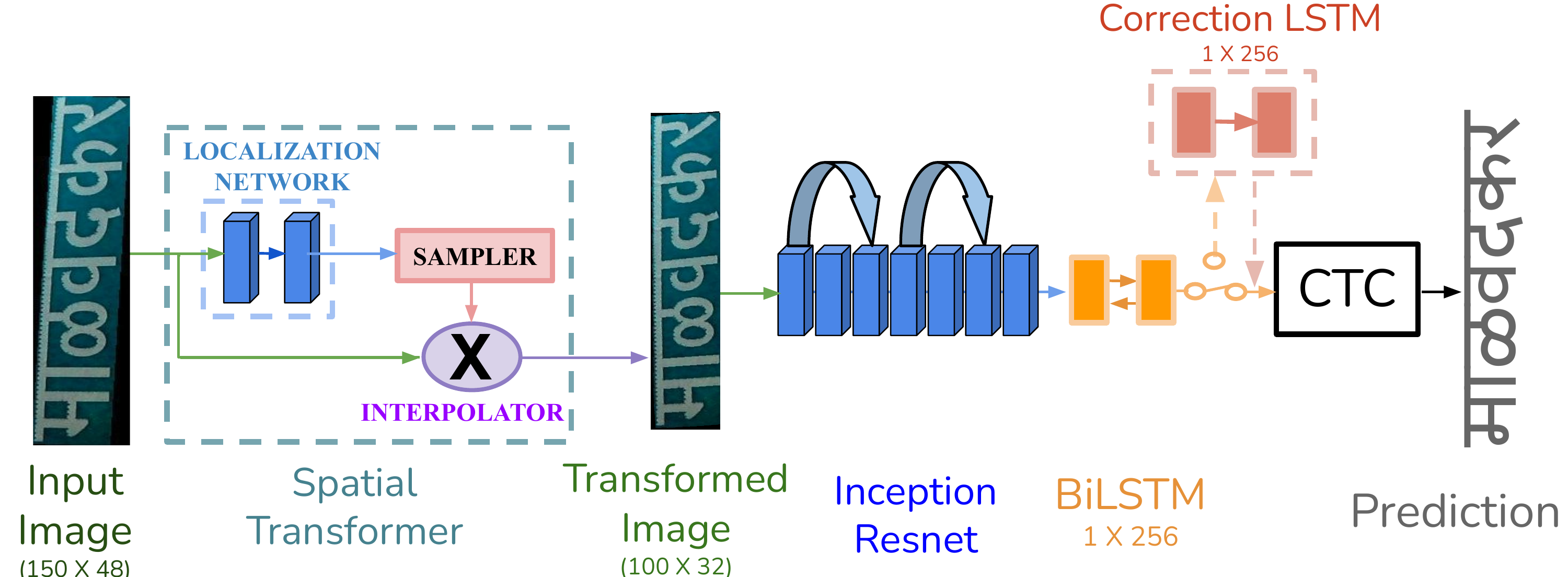}%
    \caption{Model used to train on non-Latin datasets.}
    \label{fig:asar_model}
\end{figure}
\section{Underlying Model}\label{sec:models}
We now describe the model we train for our experiments. We use STAR-Net because of its capacity to handle different image distortions~\cite{liu2016star}. It has a Spatial Transformer network, a Residue Feature Extractor, and a Connectionist Temporal Classification (CTC) layer. As shown in Fig.~\ref{fig:asar_model}, the first component consists of a spatial attention mechanism achieved via a CNN-based localisation network that helps predict affine transformation parameters to handle image distortions. The second component consists of a Convolutional Neural Network (CNN) and a Recurrent Neural Network (RNN). The CNN is inception-resnet architecture, which helps in extracting robust image features~\cite{szegedy2017inception}. The last component provides the non-parameterized supervision for text alignment. The overall end-to-end trainable model consists of $26$ convolutional layers~\cite{liu2016star}.

The input to spatial transformer module is of resolution $150\times48$. The spatial transformer outputs the image of size $100\times32$ for the next stage (Residue Feature Extractor). We train all our models on $5M$ synthetic word images as discussed in the previous section. We use the batch size of $32$ and the ADADELTA
optimizer for our experiments~\cite{article}. We train each model for $10$ epochs and test on Arabic and Devanagari word images from IIIT-ILST, MLT-17, and MLT-19 datasets. Only for the Arabic MLT-17 dataset, we fine-tune our models on training images and test them on validation images to fairly compare with Bušta et al.~\cite{buvsta2017deep}. For Devanagari, we present the additional results on the IIIT-ILST dataset by fine-tuning our best model on the MLT-19 dataset. We fine-tune all the layers of our model for the two settings mentioned above. To further improve our models, we add an LSTM layer of size $1\times256$ to the STAR-Net model, pre-trained on synthetic data. The additional layer corrects the model's bias towards the synthetic datasets, and hence we call it correction LSTM. We plug-in the correction LSTM before the CTC layer, as shown in Fig.~\ref{fig:asar_model} (top-right). After attaching the LSTM layer, we fine-tune the complete network on the real datasets. 
%

%% file: 5Results.tex
\begin{table}[t]
\resizebox{0.88\textwidth}{!}
{%
\centering
\begin{tabular}{llccrr}
\toprule

{\bf Language} & {\bf Dataset} & {\bf \# Images} & {\bf Model} & {\bf CRR}  & {\bf WRR} \\
\midrule
\multirow{5}{2.7em}{Arabic} & \multirow{5}{3.7em}{MLT-17} & \multirow{5}{2.7em}{951} & Bušta et al.~\cite{buvsta2018e2e} & 75.00 & 46.20\\
 &  &  & STAR-Net (85 Fonts) FT & 88.48 & 66.38 \\
 &  &  & STAR-Net (140 Fonts) FT & 89.17 & 68.51\\
 &  &  & STAR-Net (140 Fonts) FT & \multirow{2}{2.4em}{\bf{90.19}} & \multirow{2}{2.4em}{\bf{70.74}}\\
&  &  &  with Correction LSTM&  &  \\
\midrule
\multirow{6}{2.7em}{Devanagari} & \multirow{6}{5.7em}{IIIT-ILST} & \multirow{6}{2.7em}{1150} & Mathew et al.~\cite{mathew2017benchmarking} &  75.60 & 42.90\\
&  &  & STAR-Net (97 Fonts) & 77.44 & 43.38 \\
&  &  & STAR-Net (194 Fonts) & 77.65 & 44.27\\
&  &  & STAR-Net (194 Fonts) FT & \multirow{2}{2.24em}{79.45} & \multirow{2}{2.24em}{50.02}\\
&  &  &  on MLT-19 data &  &  \\
&  &  & STAR-Net (194 Fonts) FT & \multirow{2}{2.4em}{\bf{80.45}} & \multirow{2}{2.4em}{\bf{50.78}}\\
&  &  &  with Correction LSTM &  &  \\
\midrule
\multirow{2}{2.7em}{Arabic} & \multirow{2}{3.7em}{MLT-19} & \multirow{2}{2.7em}{4501} & STAR-Net (85 Fonts) & 71.15 & 40.05\\ 
&  &  & STAR-Net (140 Fonts) & \multirow{1}{2.4em}{\bf{75.26}} &  \multirow{1}{2.4em}{\bf{42.37}}\\%
\midrule
\multirow{2}{2.7em}{Devanagari} & \multirow{2}{3.7em}{MLT-19} & \multirow{2}{2.7em}{3766} & STAR-Net (97 Fonts)  & 84.60 & 60.83  \\ 
&  &  & STAR-Net (194 Fonts) & \multirow{1}{2.4em}{\bf{85.87}} &  \multirow{1}{2.4em}{\bf{64.55}}\\
\bottomrule
\end{tabular}
}
\caption{Results of our experiments on real datasets. FT means fine-tuned.}
\label{tab:results_real}
\end{table}
\section{Results}\label{sec:results}

Table~\ref{tab:results_real} depicts the performance of our experiments on the real datasets. For the Arabic MLT-17 dataset and Devanagari IIIT-ILST dataset, we achieve recognition rates better than Bušta et al.~\cite{buvsta2017deep} and Mathew et al.~\cite{mathew2017benchmarking}. With STAR-Net model trained on $<100$ fonts (refer Section~\ref{sec:datasets}), we achieve $13.48\%$ and $20.18\%$ gains in Character Recognition Rate (CRR) and Word Recognition Rate (WRR) for Arabic, and $1.84\%$ and $0.48\%$ improvements for Devanagari over the previous works (compare rows 1, 2 and 5, 6 in the last column of Table~\ref{tab:results_real}). The CRR and WRR further improve by training the models on the same amount of training data synthesized with $>=140$ fonts (rows 3 and 7 in the last column of Table~\ref{tab:results_real}). By fine-tuning the Devanagari model on the MLT-19 dataset, the CRR and WRR gains raise to $3.85\%$ and $7.12\%$. By adding the correction LSTM layer to the best models, we achieve the highest CRR and WRR gains of  $15.19\%$ and $24.54\%$ for Arabic, and $5.25\%$ and $7.88\%$ for Devanagari, over the previous works. The final results for the two datasets discussed above can be seen in rows 3 and 7 of the last column of Table~\ref{tab:results_real}.

As shown in Table~\ref{tab:results_real}, for the MLT-19 Arabic dataset, the model trained on $5M$ samples generated using $85$ fonts achieve the CRR of $71.15\%$ and WRR of $40.05\%$. Increasing the number of diverse fonts to $140$ gives a CRR gain of $4.11\%$ and a WRR gain of $2.32\%$. For the MLT-19 Devanagari dataset, the model trained on $5M$ samples generated using $97$ fonts achieves the CRR of $84.60\%$ and WRR of $60.83\%$. Increasing the number of fonts to $194$ gives a CRR gain of $1.27\%$ and a WRR gain of $3.72\%$. It is also interesting to note that the WRR of our models on MLT-17 Arabic and MLT-19 Devanagari datasets are very close to the WRR of the English model trained on $5M$ samples generated using $100$ fonts (refer to the yellow curve in Fig.~\ref{fig:LATIN_WRR}). It supports our claim that the number of fonts used to create the synthetic dataset plays a crucial role in improving the photo OCR models in different languages.
 
\begin{figure}[t]
    \centering
    \includegraphics[width=0.49\linewidth]{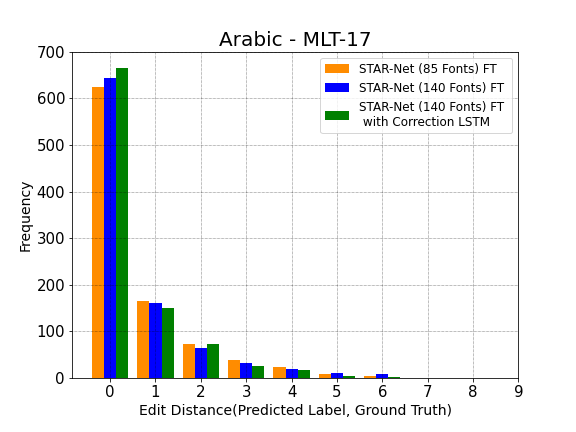}
    \includegraphics[width=0.49\linewidth]{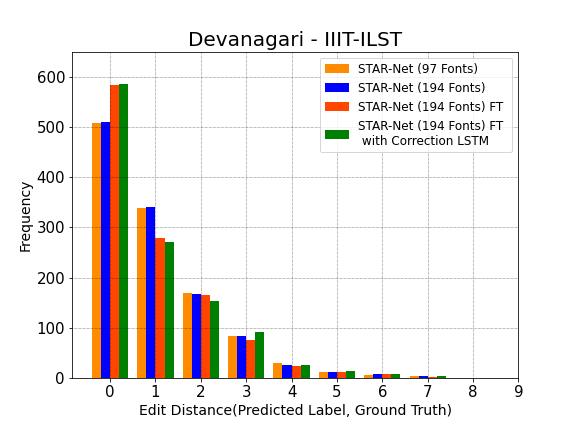}
    \includegraphics[width=0.49\linewidth]{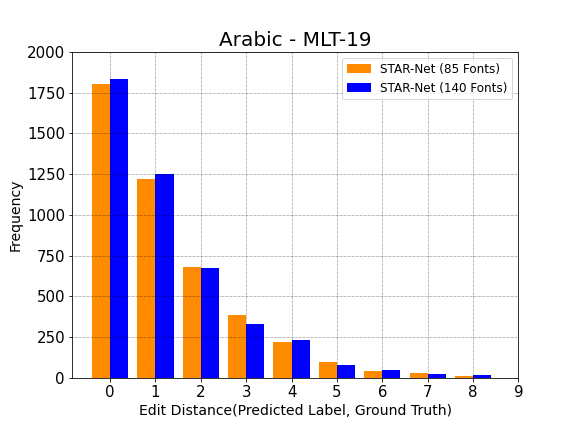}
    \includegraphics[width=0.49\linewidth]{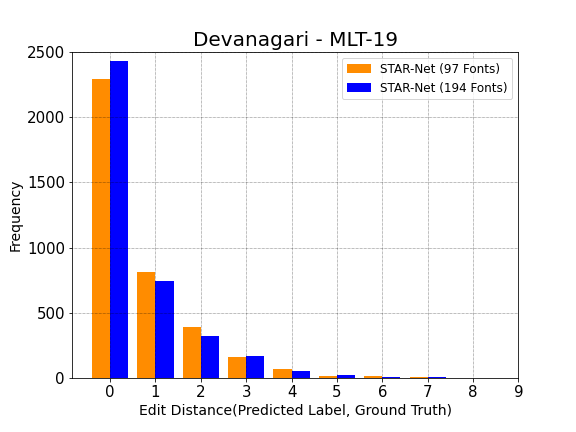}
    \caption{Histogram of correct words ($x=0$) and words with $x$ errors ($x>0$). FT represents the models fine-tuned on real datasets.}
    \label{fig:histograms}
\end{figure}

To present the overall improvements by utilizing extra fonts and correction LSTM at a higher level, we examine the histograms of edit distance between the pairs of predicted and corresponding ground truth words in Fig.~\ref{fig:histograms}. Such histograms are used in one of the previous works on OCR error corrections~\cite{rohit2017}. The bars at the edit distance of $0$ represent the words correctly predicted by the models. The subsequent bars at edit distance $n>0$ represent the number of words with $x$ erroneous characters. As it can be seen in Fig.~\ref{fig:histograms}, overall, with the increase in the number of fonts and subsequently with correction LSTM, i) the number of correct words ($x=0$) increase for each dataset, and ii) the number of incorrect words ($x>0$) reduces for many values of $x$ for the different datasets. We observe few exceptions in each histogram where the frequency of incorrect words is higher for the best model than others, e.g., at edit distance of $2$ for the Arabic MLT-17 dataset. The differences (or exceptions) show that the recognitions by different models complement each other.
\begin{figure}[h!t]
    \centering
    \includegraphics[width=0.4945\linewidth]{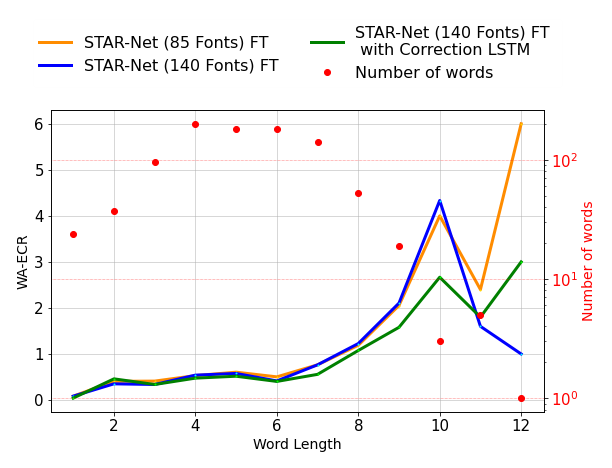}
    \includegraphics[width=0.4945\linewidth]{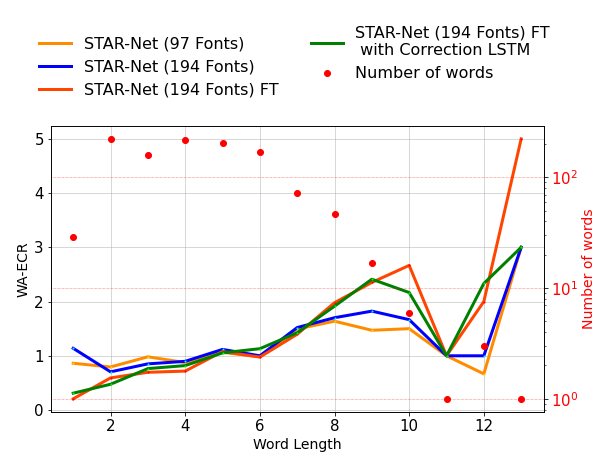}
    \includegraphics[width=0.4945\linewidth]{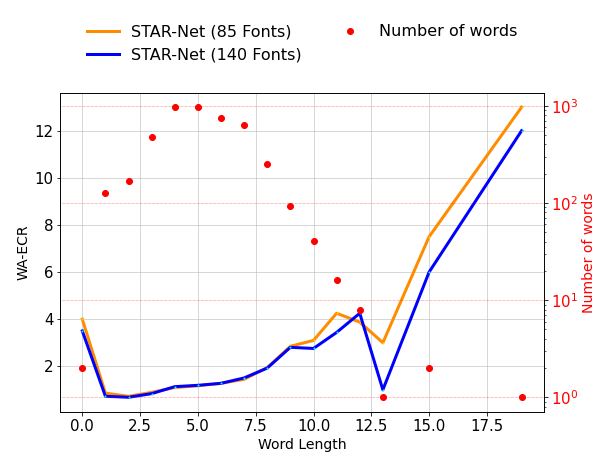}
    \includegraphics[width=0.4945\linewidth]{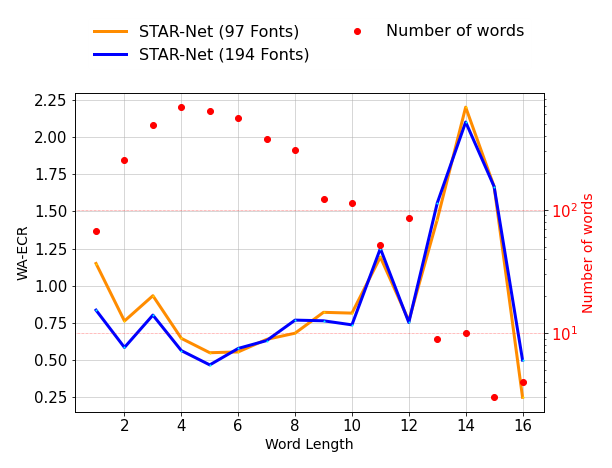}
    \caption{Clockwise from top-left: WA-ECR of our models tested on MLT-17 Arabic, IIIT-ILST Devanagari, MLT-19 Devanagari, and MLT-19 Arabic datasets.}
    \label{fig:wa_ecr}
\end{figure}

Another exciting way to compare the output of different OCR systems is Word-Averaged Erroneous Character Rate (WA-ECR), as proposed by Agam et al.~\cite{agamSanOCR2020}. The WA-ECR is the ratio of i) the number of erroneous characters in the set of all $l$-length ground truth words ($e_l$), and ii) the number of $l$-length ground truth words ($n_l$) in the test set. As shown in the red dots and the right y-axis of the plots in Fig.~\ref{fig:wa_ecr}, the frequency of words generally reduces with an increase in word length after $x = 4$. Therefore, the denominator term tends to decrease the WA-ECR for short-length words. Moreover, as the word length increases, it becomes difficult for the OCR model to predict all the characters correctly. Naturally, the WA-ECR tends to increase with the increase in word length for an OCR system. In Fig.~\ref{fig:wa_ecr}, we observe that our models trained on $>=140$ fonts (blue curves) are having lower WA-ECR across different word lengths as compared to the ones trained on $<100$ fonts (orange curves). For the IIIT-ILST dataset, the model, trained on $194$ fonts, performs poorly on the long words ($x>8$ in the top-right plot of Fig.~\ref{fig:wa_ecr}), and the correction LSTM further enhances this effect. On the contrary, we observe that the Correction LSTM reduces WA-ECR for the MLT-17 Arabic dataset for word lengths in the range $[6,11]$ (compare green and blue curves in the top-left plot). Interestingly, the WA-ECR of some of our models drops after word-length of $10$ and $14$ for the MLT-19 Arabic and MLT-19 Devanagari datasets (see blue curve in the top-left plot and the two curves in the bottom-right plot of Fig.~\ref{fig:wa_ecr}).

\begin{figure}[t]
    \centering
    \includegraphics[trim=30 72 60 128, clip,width=\linewidth]{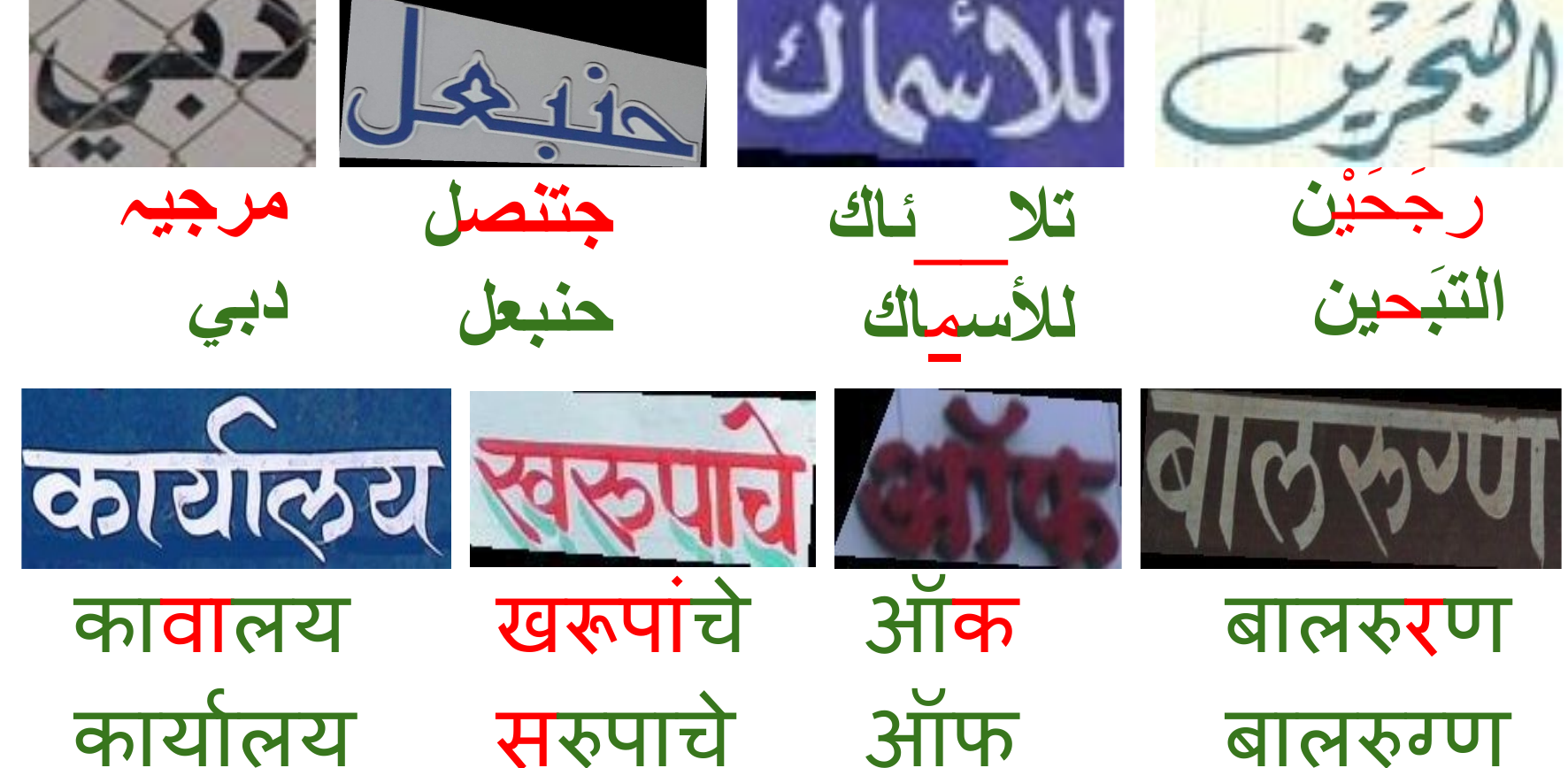}%
    \caption{Real word images in Arabic (top) and Devanagari (bottom), Below the images: predictions from i) baseline model trained on $<100$ fonts, ii) model trained on $\geq140$ fonts.  Green \& red represent correct predictions and errors.} 
    \label{fig:sample_results}
\end{figure}

In Fig.~\ref{fig:sample_results}, we present the qualitative results of our models. The green and red colors represent the predictions and errors. As shown, the models trained on over $140$ fonts perform better than the models trained on $<100$ fonts. Overall, the experiments support our claim that the diversity in fonts used to generate synthetic datasets is crucial for improving the existing non-Latin scene-text recognition systems.

%% file: 6Conclusions.tex
\section{Conclusion}~\label{sec:conclusions}
We carried out a series of controlled experiments in English to highlight the importance of font diversity and the number of synthetic examples in improving the scene-text recognition accuracy. We augmented the font set of two non-Latin scripts, Arabic and Devanagari, with new fonts obtained by region-based online search. We generated $5M$ synthetic images in two languages. Our experiments show improvements over the previous works and baselines trained on lesser fonts. We further improve our results by introducing the correction LSTM into the models to reduce the bias towards the synthetic data. Finally, we affirm that more fonts are required to improve the existing non-Latin systems. For future work in this area, we plan to employ human designers or Generative Adversarial Networks (GAN) based font generators to boost the accuracy of non-Latin scene-text recognition.